\pdfoutput=1
\documentclass[runningheads]{llncs}

\usepackage[year=2026]{eccv}

\usepackage{eccvabbrv}
\usepackage{algorithm}
\usepackage{algorithmic}
\usepackage{graphicx}
\usepackage{booktabs}
\usepackage{siunitx}
\usepackage{adjustbox}
\usepackage[table]{xcolor}
\usepackage{multirow}
\usepackage{multicol}
\usepackage{wrapfig}
\usepackage[accsupp]{axessibility}  

\usepackage[breaklinks,colorlinks,citecolor=eccvblue,linkcolor=eccvblue,urlcolor=eccvblue]{hyperref}

\usepackage{orcidlink}
\newcommand{\method}{DIAL\xspace}
\newcommand{\map}{ISGM\xspace}
\newcommand{\mapfull}{Intrinsic Spatial Grounding Map\xspace}

\begin{document}

\title{Harnessing Intrinsic Subject-Aware Attention for Controllable Multi-Subject Video Generation}

\titlerunning{Intrinsic Subject-Aware Attention for Multi-Subject Video Generation}

\author{Niange Yu\inst{1}\thanks{Corresponding author: \email{niangeyu@gmail.com}} \and
Ye Tian\inst{2} \and
Biaolong Chen\inst{1} \and
Miao Lu\inst{1} \and
Aixi Zhang\inst{1} \and
Hao Jiang\inst{1} \and
Yunhai Tong\inst{2} \and
Pipei Huang\inst{1}}

\authorrunning{N.~Yu et al.}

\institute{Alibaba Group \and
Peking University}

\maketitle

\begin{abstract}
Multi-subject video generation faces two key challenges: uncontrollable fidelity strength and potential semantic drift. We address these by analyzing the internal mechanisms of Diffusion Transformers (DiTs). We found that certain attention blocks naturally form an Intrinsic Spatial Grounding Map (\map) that precisely locates reference subjects. 
Building on this insight, we propose Dual-phase Intrinsic Attention Leveraging (\method), a framework that uses these internal signals for both training and inference. In low-noise stages, we use \map to guide the attention mechanism, allowing precise control over fidelity strength during inference without retraining. In high-noise stages, we use these same maps to automatically build preference pairs at no additional cost for Reinforcement Learning (RL). This RL procedure effectively anchors the model's attention to reference subjects and mitigates semantic drift. 
Extensive experiments show that \method significantly outperforms baseline models on the OpenS2V-Eval benchmark, consistently improving identity consistency and enabling controllable fidelity strength.

\end{abstract}

\section{Introduction}
\label{sec:intro}
Recent advances in diffusion models~\cite{ho2020denoising,peebles2023scalable} enables significant breakthroughs in video generation~\cite{wan2025wan,yang2024cogvideox,kong2412hunyuanvideo,polyak2024movie}, delivering realistic and temporally coherent results across text-to-video (T2V)~\cite{hacohen2024ltx,wan2025wan} and image-to-video (I2V)~\cite{blattmann2023stable} tasks. Despite these successes, general-purpose models still struggle to provide \textbf{fine-grained, user-centric controllability}, especially when users require multiple personalized subjects to remain visually consistent in a dynamic sequence. 

This trend has pushed the field toward a more demanding frontier: Subject-to-Video (S2V) generation~\cite{phantom,li2025bindweave,zhang2025kaleido}. Unlike T2V, S2V requires the model to strictly adhere to the visual identities provided by \textbf{one or multiple reference images} while accurately executing complex motion instructions. This capability is critical for high-value applications such as personalized storytelling, virtual try-on, and cinematic production.

\begin{figure}[tb]
    \centering
    \includegraphics[width=1\linewidth]{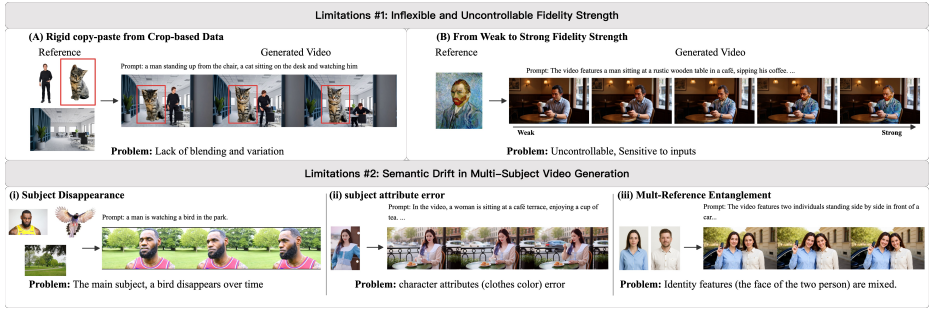}
    \caption{\textbf{Limitations} for current subject-to-video generation methods: (Top) Inflexible fidelity strength: existing models either produce unnatural ``copy-paste'' effects due to training data bias (A) or exhibit uncontrollable sensitivity to inputs (B). (Bottom) Semantic drift: models frequently suffer from (i) subject disappearance over time, (ii) attribute errors (e.g., clothing color drift), and (iii) multi-reference entanglement where identities from different subjects are mixed.}
    \label{fig:teaser1}
\end{figure}

While existing approaches have tackled consistency via high-quality dataset curation~\cite{phantom}, Vision-Language Model (VLM) feature extraction \cite{li2025bindweave}, and novel model architectures \cite{zhang2025kaleido}, two critical limitations remain largely unexplored. First, the fidelity strength in most models is \textbf{inflexible and uncontrollable}. Typically, consistency strength is dictated solely by the training data. For example, crop-based paired data often induces a rigid copy-paste effect, while cross-domain data yields weaker fidelity. As a result, users cannot dynamically adjust the consistency strength to suit different scenarios at inference time. Second, existing models often suffer from \textbf{semantic drift}, where the generated content deviates from the reference-conditioned semantics. This drift typically appears as (i) subject disappearance, (ii) attribute mismatch, and (iii) multi-reference entanglement, where identities or attributes across multiple references are mixed or swapped, as is shown in \cref{fig:teaser1}.

To address these limitations, we hypothesize that both uncontrollable fidelity and semantic drift are rooted in \textbf{misaligned attention} to reference subjects during denoising. We therefore investigate the internal mechanics of Diffusion Transformers (DiTs) with a layer-wise analysis of cross-attention. We obtain subject masks from a semantic segmentation pipeline and use them as ground truth, and measure the spatial correlation between these masks and the attention weights at each layer and timestep. 

\begin{figure}[tb]
    \centering
    \includegraphics[width=1\linewidth]{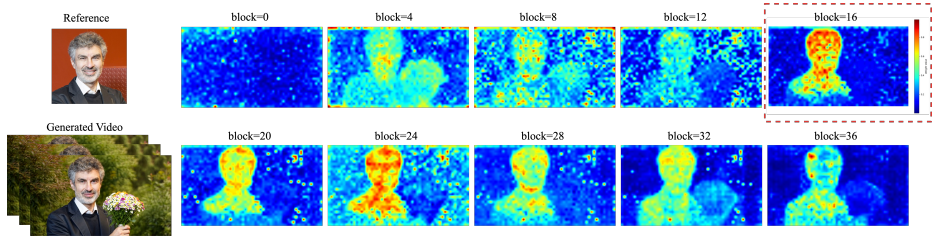}
    \caption{Visualization of the attention peaking phenomenon and Intrinsic Spatial Grounding Maps (ISGM) in Kaleido~\cite{zhang2025kaleido} Baseline. Layer-wise analysis reveals that while most attention blocks produce diffuse noise, a certain block consistently exhibits the most spatially concentrated grounding to the reference subject. }
    \label{fig:teaser_attention}
\end{figure}

This analysis reveals a surprising pattern(\cref{fig:teaser_attention}): Certain attention block(s) consistently assigns \textbf{the strongest and most spatially concentrated} attention to the reference subjects, yielding markedly higher grounding accuracy than other layers. We term these \textit{peaking} internal signals the \mapfull~(\map). The \map precisely localizes reference subjects in the video latent space, providing an intrinsic guidance signal that can be leveraged to modulate and strengthen subject consistency throughout the denoising process.

Inspired by this finding, we propose \textbf{Dual-phase Intrinsic Attention Leveraging (\method)}, which exploits \map for both training-free inference control and training-time alignment.

In the \textbf{late denoising stages}, \map is sharp and spatially reliable. We extract \map from the identified blocks, binarize it into subject-specific spatial guides, and use them to inject a \textbf{controllable attention bias} into all attention blocks during denoising. This mechanism is training-free and also supports independent, adjustable guidance for each reference subject in multi-subject scenarios, leading to large gains in identity consistency.

In the \textbf{early denoising stages}, attention is noisier and \map is not suitable for direct supervision. To better exploit its signal, we introduce a \textbf{zero-cost preference construction} pipeline for Reinforcement Learning (RL) fine-tuning. Specifically, we score generated denoising trajectories by measuring how well their \map aligns with the ground-truth masks. These scores allow us to automatically build preference pairs without manual annotation, and we use them to fine-tune the model to anchor attention to reference subjects. This procedure stabilizes reference grounding and mitigates potential semantic drift.

Applied to strong subject-to-video baselines~\cite{zhang2025kaleido, phantom}, \method delivers consistent improvements in both quantitative and qualitative evaluations, and achieves \textbf{state-of-the-art} performance on the OpenS2V-Eval~\cite{yuan2025opens2v} benchmark. Further analyses and ablations show that \method enables free and fine-grained fidelity control at inference time without sacrificing overall visual quality. Moreover, it substantially reduces semantic drift and leads to more reliable subject grounding. Our contributions are as follows:
\begin{enumerate}
\item We uncover a subset of \emph{high-quality} attention blocks in DiT architectures whose attention patterns form an \textbf{Intrinsic Spatial Grounding Map (ISGM)} that precisely localizes reference subjects.
\item We propose a \textbf{training-free}, inference-time guided attention mechanism that enables \textbf{precise, region-adaptive, and user-controllable} fidelity enhancement for multi-subject generation.
\item We introduce a \textbf{zero-cost} preference pair construction strategy and leverage reinforcement learning to effectively anchor reference grounding and mitigate semantic drift.
\item By combining these components, our unified framework \method achieves \textbf{state-of-the-art} performance on the \textbf{OpenS2V-Eval} benchmark~\cite{yuan2025opens2v}.
\end{enumerate}

\section{Related Work}

\subsection{Subject-to-Video Generation}
Subject-to-video (S2V) generation has evolved from computationally expensive per-subject optimization or fine-tuning~\cite{wang2024customvideo, chen2024disenstudio} to efficient end-to-end adapters that enable zero-shot identity preservation~\cite{phantom, jiang2025vace, li2025bindweave}. Representative frameworks such as Phantom~\cite{phantom} and VACE~\cite{jiang2025vace} utilize joint injection and unified context adapters to facilitate multimodal alignment. More recent models like HunyuanCustom~\cite{hu2025hunyuancustom}, MAGREF~\cite{deng2025magref}, and PolyVivid~\cite{hu2025polyvivid} have further refined identity injection through region-aware masking and 3D-RoPE. Notably, Kaleido~\cite{zhang2025kaleido} improves reference grounding via Reference Rotary Positional Encoding (R-RoPE) and curated data construction. However, these models still frequently suffer from semantic drift—such as subject disappearance or attribute entanglement—and lack a mechanism for user-controllable fidelity strength, as the consistency level is typically fixed by the training data bias~\cite{phantom}.

\subsection{Attention Analysis in Diffusion Models}
Attention maps are widely utilized as diagnostic and control interfaces in diffusion models. In U-Net architectures, cross-attention maps facilitate grounding-aware editing~\cite{hertz2022prompt} and subject emphasis~\cite{chefer2023attend}, while self-attention serves as a structural cue for training-free guidance in methods like SAG~\cite{hong2023improving} and PAG~\cite{ahn2024self}. Recently, these techniques have been extended to Diffusion Transformers (DiTs). DiTCtrl~\cite{cai2025ditctrl} explores attention control in MM-DiT for coherent video generation, and DiT4Edit~\cite{feng2025dit4edit} adapts these controls for image editing. Despite this progress, existing literature rarely characterizes which specific internal DiT blocks provide the most reliable \mapfull (\map) for multi-subject synthesis. \method fills this gap by identifying privileged attention layers that naturally act as grounding modules.

\subsection{Reinforcement Learning for Diffusion Models}
Reinforcement learning (RL) has become a practical tool for aligning diffusion models with downstream objectives. Early methods like DDPO~\cite{black2023training} optimized models via policy gradients, while modern approaches favor preference-based alignment. Diffusion-DPO~\cite{wallace2024diffusion} adapted Direct Preference Optimization to the diffusion paradigm, and video-specific variants like VideoDPO~\cite{liu2025videodpo} and DanceGRPO~\cite{xue2025dancegrpo} have scaled these techniques to temporal generation. However, these methods typically depend on expensive human feedback or external reward models. In contrast, \method introduces a zero-cost preference construction pipeline that scores denoising trajectories based on the spatial alignment between the \map and ground-truth subject masks, enabling automated anchoring of subject semantics.

\section{Method}
\label{sec:isgm_identification}
\begin{figure}[tb]
    \centering
    \includegraphics[width=1\linewidth]{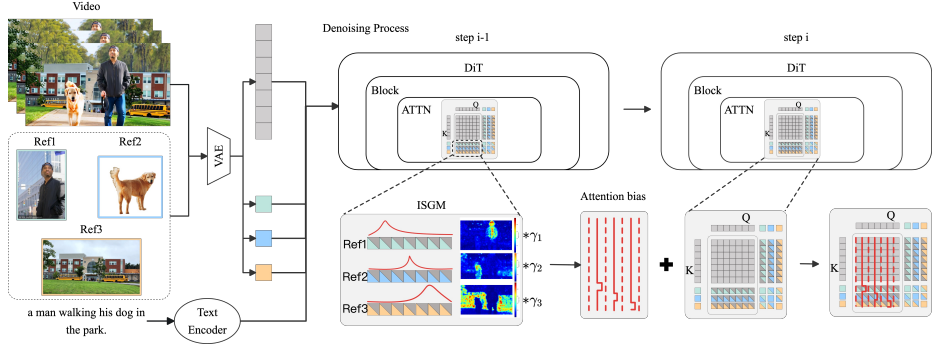}
    \caption{\textbf{Phase I: Attention Guidance via ISGM.} This training-free mechanism extracts the Intrinsic Spatial Grounding Map (ISGM) from step $i-1$ to guide the self-attention (ATTN) in step $i$. Each subject has an independent, spatially-aligned ISGM used to construct an attention bias matrix. The guidance strength ($\gamma_k$) can be adjusted independently for each subject, enabling precise identity fidelity control without requiring model updates.}
    \label{fig:method2}
\end{figure}

\begin{figure}[tb]
    \centering
    \includegraphics[width=1\linewidth]{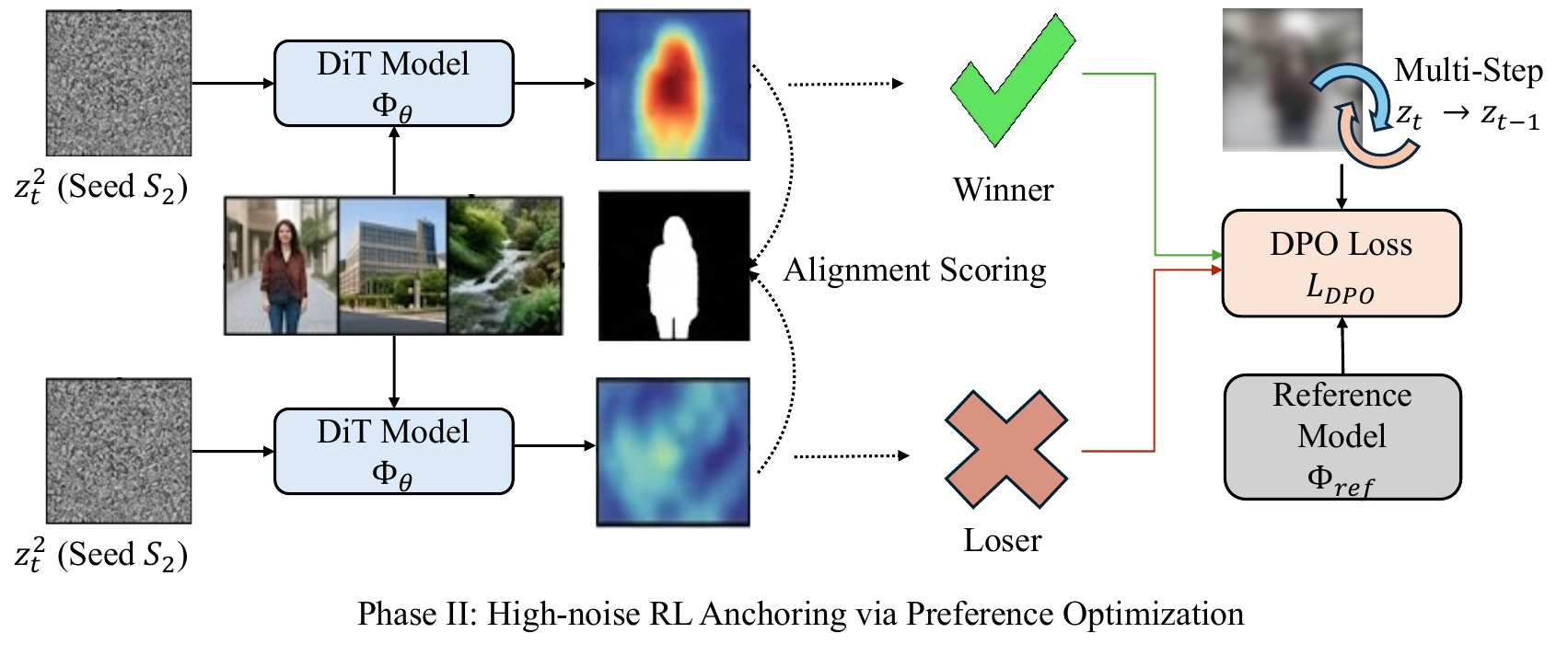}
    \caption{\textbf{Phase II: Dual-Seed Preference Optimization.} We construct preference pairs by evaluating the spatial alignment between the subject mask $\mathbf{M}_k$ and ISGMs generated from distinct random seeds. Then we use DPO to optimize the model policy, enabling proactive anchoring of subject semantics and mitigation of drift during the high-noise denoising phase.}
    \label{fig:method}
\end{figure}

\subsection{Preliminaries}

\subsubsection{Video Diffusion Models}
We adopt a text-conditioned Diffusion Transformer (DiT) architecture \cite{peebles2023scalable} for latent video generation. Let $\mathbf{x}_0=\{\mathbf{x}_0^{(f)}\}_{f=1}^{F}$ denote a video clip with $F$ frames. A spatio-temporal VAE encoder $E(\cdot)$ maps the video into a compact latent representation $\mathbf{z}_0 = E(\mathbf{x}_0) \in \mathbb{R}^{T \times C \times H \times W}$ \cite{wan2025wan}. Following the rectified flow matching framework \cite{liu2022flow, wan2025wan}, generation is modeled as an interpolation between the data latent $\mathbf{z}_0$ and Gaussian noise $\mathbf{z}_1 \sim \mathcal{N}(\mathbf{0}, \mathbf{I})$. For $t \sim \mathcal{U}(0, 1)$, the noisy latent is formed as:
\begin{equation}
\mathbf{z}_t = (1-t)\mathbf{z}_0 + t\mathbf{z}_1,
\end{equation}
with the target velocity defined as $\mathbf{v}_t = \frac{d\mathbf{z}_t}{dt} = \mathbf{z}_1 - \mathbf{z}_0$. A DiT-based network predicts the velocity $\mathbf{v}_\theta(\mathbf{z}_t, t, \mathbf{c})$ conditioned on text tokens $\mathbf{c}$, and is optimized using the flow-matching objective:
\begin{equation}
\mathcal{L}_{\mathrm{fm}} = \mathbb{E}_{t, \mathbf{z}_0, \mathbf{z}_1} \left[ \left\| \mathbf{v}_t - \mathbf{v}_\theta(\mathbf{z}_t, t, \mathbf{c}) \right\|_2^2 \right].
\end{equation}
During inference, the video is generated by numerically solving the ODE $d\mathbf{z}_t/dt = \mathbf{v}_\theta(\mathbf{z}_t, t, \mathbf{c})$ from $t=1$ to $t=0$, followed by VAE decoding $\hat{\mathbf{x}}_0 = D(\mathbf{z}_0)$ \cite{wan2025wan}.

\subsubsection{Subject-to-Video Generation}
Subject-to-video (S2V) generation aims to synthesize a video $\hat{\mathbf{x}}_0$ that aligns with a text prompt $\mathbf{y}$ while maintaining the visual identity of a set of reference subjects $\mathcal{R} = \{\mathbf{I}_k\}_{k=1}^{K}$ \cite{phantom, zhang2025kaleido}. 

Following current state-of-the-art backbones \cite{wan2025wan, zhang2025kaleido}, reference information is injected through intrinsic latent-token interaction. Each reference image $\mathbf{I}_k$ is mapped to a spatial latent $\mathbf{z}^{(k)}_{\mathrm{img}} = E(\mathbf{I}_k)$. At each denoising step $t$, the video latent $\mathbf{z}_t$ and all reference latents are concatenated:
\begin{equation}
\tilde{\mathbf{z}}_t = \mathrm{Concat}_c \left( \mathbf{z}^{(1)}_{\mathrm{img}}, \dots, \mathbf{z}^{(K)}_{\mathrm{img}},\mathbf{z}_t \right).
\end{equation}
The resulting tensor is patchified and linearly projected to form the initial token sequence $\mathbf{H}^{(0)} = \mathrm{PatchEmbed}(\tilde{\mathbf{z}}_t)$. In this unified sequence, video and reference tokens coexist, enabling the DiT's self-attention layers to perform joint spatio-temporal modeling and cross-modal identity transfer \cite{zhang2025kaleido}. Textual conditions are typically incorporated via cross-attention. In our framework, we primarily focus on manipulating these spatio-temporal self-attention blocks to achieve controllable fidelity and mitigate semantic drift.

\subsection{Identification of Intrinsic Spatial Grounding Maps (\map)}
We begin with a practical observation on current open-source S2V systems~\cite{zhang2025kaleido, phantom} as shown in \cref{fig:teaser1}: despite strong overall quality, they frequently exhibit semantic drift in real usage. In these failure cases, the reference subjects are missing, attributes deviate, or multiple references become entangled. These errors suggest a mismatch between \emph{what the references specify} and \emph{how the model actually uses them} during denoising. Intuitively, this points to a failure in the interaction between the noisy video latent and reference-image latents, which in DiT backbones is primarily mediated by self-attention over the joint token sequence.
To resolve this, we conducted a layer-wise investigation of the attention blocks to locate the source of this misalignment.

\paragraph{Token Layout.}
At denoising step $i$, the DiT processes a token sequence $\mathbf{H}_i = [\mathbf{H}^{\text{ref}}_{i,1}; \ldots; \mathbf{H}^{\text{ref}}_{i,K}; \mathbf{H}^{\text{vid}}_i]$. Here, $\mathbf{H}^{\text{ref}}_{i,k}$ represents tokens from the $k$-th reference latent and $\mathbf{H}^{\text{vid}}_i$ denotes tokens from the noisy video latent. Given $N_k$ as the token length of $\mathbf{H}^{\text{ref}}_{i,k}$ and $N_b$ as the length of $\mathbf{H}^{\text{vid}}_i$, the total sequence length is $N = \sum_{k=1}^{K} N_k + N_b$.

\paragraph{Extracting per-subject Attention Maps.}
For each self-attention module, let queries and keys be $\mathbf{Q} \in \mathbb{R}^{h \times N \times d}$ and $\mathbf{K} \in \mathbb{R}^{h \times N \times d}$. We measure the attention from video tokens to each reference subject by computing:
\begin{equation}
\mathbf{A} = \frac{1}{h} \sum_{m=1}^{h} \mathrm{softmax} \left( \frac{\mathbf{Q}^{(m)}_{\text{vid}}(\mathbf{K}^{(m)})^\top}{\sqrt{d}} \right) \in \mathbb{R}^{N_b \times N}
\end{equation}
where $\mathbf{Q}^{(m)}_{\text{vid}}$ corresponds to queries from the video tokens. The per-subject attention score is then defined as:
\begin{equation}
\mathbf{s}_k = \sum_{j \in \mathcal{I}_k} \mathbf{A}_{:,j} \in \mathbb{R}^{N_b}, \quad \mathcal{I}_k = \left[ \sum_{\ell < k} N_\ell, \sum_{\ell \le k} N_\ell \right)
\end{equation}
We aggregate these into a joint score matrix $\mathbf{S} = [\mathbf{s}_1, \ldots, \mathbf{s}_K] \in \mathbb{R}^{N_b \times K}$.

\begin{wrapfigure}{r}{0.42\linewidth}
  \centering
  \includegraphics[width=\linewidth]{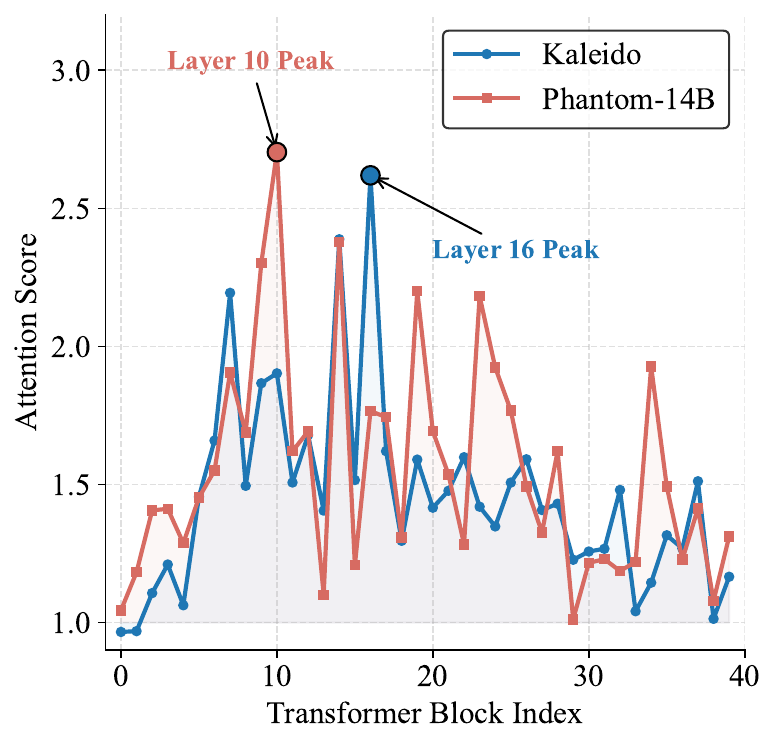}
  \caption{Attention peaking across blocks.}
  \label{fig:isgm_peak}
\end{wrapfigure}
\paragraph{The Peaking Phenomenon.}

We evaluate $\mathbf{S}$ for each attention block and identify the block that yields the highest grounding accuracy. Our analysis is shown in ~\cref{fig:isgm_peak}. This reveals a consistent \textbf{peaking phenomenon} across diverse prompts. Among the numerous attention blocks, a specific block yields the highest interaction score. The attention map in the specific block is highly concentrated and aligns precisely with the subject regions. We define this refined signal as the \mapfull~(\map). The discovery of these intrinsic maps suggests that the DiT backbone contains a specialized grounding layer for subject-to-video information transfer. This layer serves as the foundation for our Dual-phase Intrinsic Attention Leveraging(\method).

\subsection{\method Phase I: Low-noise Inference Guidance}


Building on the discovery of ISGMs, we first utilize this intrinsic attention signal to implement training-free fidelity control. Our observations indicate that during the later stages of the denoising process, typically the final 80\% of timesteps, the attention maps for individual subjects become highly structured and spatially consistent. These maps serve as reliable spatial anchors to guide the synthesis of fine-grained details in subsequent steps.

Let $\mathbf{S}_{i-1}$ be the extracted \map from step $i\!-\!1$.
We normalize it to obtain a soft spatial guide
$\tilde{\mathbf{S}}_{i-1}(:,k)=\mathbf{S}_{i-1}(:,k)/\max(\mathbf{S}_{i-1}(:,k))$.
Optionally, a binary guide can be obtained by thresholding.
At step $i$, we modulate all self-attention modules in the DiT backbone by injecting a controllable bias matrix $\mathbf{M}_i$ into the attention logits. This bias specifically targets the interaction between video queries and reference subject keys:
\begin{equation}
\label{eq:mask}
\mathbf{M}_i(q, j) = 
\begin{cases} 
\log(\gamma_k) \cdot \tilde{\mathbf{S}}_{i-1}(q, k), & q \in \mathcal{I}_{\text{vid}}, j \in \mathcal{I}_k, \\
0, & \text{otherwise}
\end{cases}
\end{equation}
where $\gamma_k>0$ is a user-specified fidelity scale for subject $k$,
$\mathcal{I}_{\text{vid}}=[\sum_{k=1}^{K}N_k,\ N)$ denotes the index range of video tokens,
and $\mathcal{I}_k$ denotes the key range of the $k$-th reference as above.

For an attention module at step $i$,
the biased attention is computed as
\begin{equation}
\mathrm{Attn}_i(\mathbf{Q}_i,\mathbf{K}_i,\mathbf{V}_i)=
\mathrm{softmax}\!\left(\frac{\mathbf{Q}_i\mathbf{K}_i^\top}{\sqrt{d}}+\mathbf{M}_i\right)\mathbf{V}_i.
\label{eq:phase1_attn}
\end{equation}
Intuitively, $\mathbf{M}_i$ selectively increases the attention paid by guided video regions to the corresponding reference subject,
enabling \emph{independent} and \emph{continuous} fidelity control in multi-subject generation.
This mechanism, as is shown in \cref{fig:method2}, is training-free and requires no model updates; it only reuses \map from the previous denoising step and adds a lightweight log-scale bias. This process is illustrated in \cref{alg:guidance} and \cref{fig:method2}.

\begin{algorithm}[t]
\caption{\method Phase I: Low-noise Inference Guidance}
\label{alg:guidance}
\begin{algorithmic}[1]
\REQUIRE Pre-trained DiT model $\Phi$, Reference subjects $R=\{R_1, \dots, R_K\}$, Text prompt $T$, Fidelity scales $\Gamma=\{\gamma_1, \dots, \gamma_K\}$, Denoising steps $N$.
\ENSURE Generated video latent $\mathbf{z}_0$.
\STATE Initialize $\mathbf{z}_N \sim \mathcal{N}(0, I)$; 
\STATE $\mathbf{S}^{(N+1)} \leftarrow \text{None}$; 
\FOR{$i = N$ \TO $1$}
    \STATE // \textit{Step 1: Fidelity Bias Construction}
    \IF{$\mathbf{S}^{(i+1)}$ is not None \AND $i \le 0.8N$}
        \STATE Normalize $\mathbf{S}^{(i+1)}$ to obtain spatial guide $\tilde{\mathbf{S}}^{(i+1)}$;
        \STATE Construct Bias Matrix $\mathbf{M}_i$ using $\tilde{\mathbf{S}}^{(i+1)}$ and $\Gamma$; \COMMENT{\cref{eq:mask}}
        \STATE Inject $\mathbf{M}_i$ into DiT attention modules;
    \ELSE
        \STATE Set $\mathbf{M}_i = \mathbf{0}$ (Disable guidance);
    \ENDIF
    
    \STATE // \textit{Step 2: Denoising and Map Extraction}
    \STATE $[v_\theta(\mathbf{z}_i), \mathbf{A}^{(i)}] \leftarrow \Phi(\mathbf{z}_i, T, R, \mathbf{M}_i)$; 
    \STATE $\mathbf{z}_{i-1} \leftarrow \text{Denoise}(\mathbf{z}_i, v_\theta(\mathbf{z}_i))$; 
    
    \STATE // \textit{Step 3: \map Update for Next Step}
    \STATE $\mathbf{S}^{(i)} \leftarrow \text{Extract}(\mathbf{A}^{(i)}, R)$; 
\ENDFOR
\RETURN $\mathbf{z}_0$;
\end{algorithmic}
\end{algorithm}
\subsection{\method Phase II: High-noise Anchoring via Preference Optimization}
\label{sec:rl_anchoring}

While the inference-time guidance in Phase I effectively enhances subject fidelity, its performance relies heavily on the clarity of the extracted \map. Our investigations reveal that this is insufficient during the high-noise regime, typically the initial 20\% of denoising steps. In this stage, the intrinsic attention signals are highly stochastic and diffused, failing to provide the precise spatial boundaries required for deterministic bias injection. Relying solely on inference guidance during high-noise stages often leads to unstable spatial layouts and irreversible semantic drift.

Since early denoising largely determines global layout and attribute grounding, we introduce a learning-based anchoring mechanism for this regime. Although pixel-wise supervision is unreliable at high noise, we can still \emph{rank} denoising trajectories by how well their intrinsic maps align with reference subjects. We therefore cast high-noise alignment as a \textbf{preference optimization} problem in latent space and optimize the model toward trajectories with better grounding.

\paragraph{Zero-cost Preference Construction.} 
Inspired by this, we leverage the intrinsic grounding capability of \map~to automatically construct preference pairs without human annotation. For a training sample $(V, T, R)$, we first obtain the ground-truth segmentation masks $\{\mathbf{M}_k\}_{k=1}^K$ for each reference subject from the video $V$. We then generate two independent attention trajectories using different random seeds $S_1$ and $S_2$. We define a \textbf{Preference Score} $P$ based on the spatial overlap between the extracted \map~and the ground-truth mask:
\begin{equation}
P = \sum_{k=1}^K \frac{\sum (\mathbf{M}_k \odot \text{ISGM}_k)}{\sum \text{ISGM}_k}
\label{eq:pref_score}
\end{equation}
where $\odot$ denotes the element-wise product. $P$ quantifies the model's self-consistency in grounding reference subjects within the latent space. For each sample, the trajectory with the higher $P$ is designated as the winner $v^w$, while the other is the loser $v^l$.

\paragraph{Multi-step DPO Training.} 
To ensure stable semantic anchoring, we employ a multi-step Direct Preference Optimization (DPO) strategy. Unlike standard DPO that often relies on single-step noise prediction, our approach utilizes $T_{\mathrm{dpo}}$ steps to obtain a more accurate evaluation of the attention layout. For each iteration $i \in \{1, \dots, T_{\mathrm{dpo}}\}$, we compute the noise prediction $v_\theta$ and the reference model prediction $v_{ref}$ (obtained by disabling adapters). The normalized DPO loss is defined as:
\begin{equation}
\mathcal{L}_{DPO} = -\mathbb{E} \left[ \log \sigma \left( \frac{(\mathcal{L}_{ref}^w - \mathcal{L}_{ref}^l) - (\mathcal{L}_{train}^w - \mathcal{L}_{train}^l)}{\tau} \right) \right]
\label{eq:dpo_loss}
\end{equation}
where $\mathcal{L} = \|v_\theta - v_{target}\|^2$ is the mean squared error of the noise prediction, and $\tau$ is a dynamic scale factor derived from the running mean of the loss differences to maintain numerical stability. This procedure effectively anchors the model's spatial attention to the reference subjects, decoupling identity maintenance from specific prompt structures. This method is further detailed in \cref{alg:loss} and shown in \cref{fig:method}.

\begin{algorithm}[t]
\caption{\method Phase II: High-noise Anchoring via Preference Optimization}
\label{alg:loss}
\begin{algorithmic}[1]
\REQUIRE Pipeline $\Phi$, training sample $(V, T, R)$, Subject masks $\{\mathbf{M}_k\}$, DPO iterations $T_\mathrm{dpo}$, beta $\beta$.
\ENSURE Updated LoRA weights for DiT.
\STATE Sample $t \in [t_{min}, t_{max}]$ within high-noise boundary (e.g., $t \in [0, 0.2N]$);
\STATE Initialize two noisy latents $\mathbf{z}_t^1, \mathbf{z}_t^2$ with seeds $S_1, S_2$;
\FOR{$i = 1$ \TO $T_\mathrm{dpo}$}
    \STATE $v_\theta^1, A^1 \leftarrow \Phi(\mathbf{z}_t^1, T, R)$; $v_\theta^2, A^2 \leftarrow \Phi(\mathbf{z}_t^2, T, R)$;
    \STATE Compute preference scores $P_1, P_2$ using $\{\mathbf{M}_k\}$ via Eq. \ref{eq:pref_score};
    \STATE Identify winner $v^w$ and loser $v^l$ based on $\max(P_1, P_2)$;
    \STATE $v_{ref}^w, v_{ref}^l \leftarrow \Phi_{ref}(\mathbf{z}_t^{w,l}, T, R)$ (with adapters disabled);
    \STATE Compute $\mathcal{L}_{DPO}$ using normalized logits via Eq. \ref{eq:dpo_loss};
    \STATE Backpropagate and update $\theta$;
    \STATE Update $\mathbf{z}_t^{1,2}$ for the next denoising step; 
\ENDFOR
\end{algorithmic}
\end{algorithm}

\section{Experiments}
\subsection{Experimental Settings}
\label{sec:exp_settings}

\subsubsection{Benchmark and Metrics.}
We evaluate on OpenS2V-Eval~\cite{yuan2025opens2v} and follow its official protocol.
The benchmark includes 180 prompts from seven categories, covering single-subject cases (face, body, entity) as well as multi-subject and human--entity interactions.
We report the automated metrics provided by the protocol, where higher is better.
These include \textbf{Aesthetics}~\cite{improved-aesthetic-predictor} for visual appeal,
\textbf{MotionSmoothness} and \textbf{MotionAmplitude}~\cite{opencv} for temporal dynamics,
and \textbf{FaceSim}~\cite{consisid} for identity preservation.
We also report OpenS2V-Eval's benchmark-specific scores, including \textbf{NexusScore} for subject consistency,
\textbf{NaturalScore} for overall naturalness, and \textbf{GmeScore} for text--video relevance~\cite{yuan2025opens2v}.

\subsubsection{Implementation Details.}
We apply \method\ to two representative open-source S2V baselines, Phantom-14B~\cite{phantom} and Kaleido-14B~\cite{zhang2025kaleido}.
Both models are fine-tuned from the DiT-based foundation model Wan2.1-T2V-14B~\cite{wan2025wan}.
For Phase~II preference learning, we construct a training set of approximately 14K samples from Phantom-Data.
For Phase~I inference-time guidance, we use a single fidelity scale $\gamma$ shared by all subjects and report results with $\gamma\in\{0, 2,4,8, 16\}$. Here $\gamma=0$ corresponds to the baseline that disables Phase~I guidance and uses only the Phase~II preference-optimized model. More details can be found in Appendix~\ref{app:exp}.

\subsubsection{Baselines.}
We compare \method\ with recent open-sourced S2V approaches, including baseline Phantom-14B~\cite{phantom}, Kaleido~\cite{zhang2025kaleido}, along with VACE~\cite{jiang2025vace}, SkyReels-A2~\cite{fei2025skyreels},
MAGREF~\cite{deng2025magref}, BindWeave~\cite{li2025bindweave} and most recent SkyReels-v3~\cite{li2026skyreels}.

\subsection{Main Results}
\begin{table}[!t]
    \centering
    \footnotesize
    \caption{
        \textbf{Main Results on OpenS2V-Eval.} We evaluate \method~across various DiT-based models. 
        Our method, under each base model, aims to show the performance gain across different fidelity control strengths ($\gamma$). 
        Metrics: \textit{Aes.} (Aesthetics), \textit{Smth.} (MotionSmoothness), \textit{Amp.} (MotionAmplitude), \textit{Face} (FaceSim), \textit{Gme.} (GmeScore), \textit{Nex.} (NexusScore), and \textit{Nat.} (NaturalScore). 
        All values are percentages (\%). \textbf{Bold} marks the best and \underline{underline} the second best value within each block.
    }
    \label{baselines_comparison}
    \begin{tabular*}{\linewidth}{@{\extracolsep{\fill}} l c c cc ccc c}
        \toprule
        \multirow{2}{*}{Method} & Total & Quality & \multicolumn{2}{c}{Dynamics} & \multicolumn{3}{c}{Consistency} & Realism \\
        \cmidrule{2-2} \cmidrule{3-3} \cmidrule{4-5} \cmidrule{6-8} \cmidrule{9-9}
        & Score$\uparrow$ & Aes.$\uparrow$ & Smth.$\uparrow$ & Amp.$\uparrow$ & Face$\uparrow$ & Gme.$\uparrow$ & Nex.$\uparrow$ & Nat.$\uparrow$ \\
        \midrule

        \rowcolor{gray!10} \multicolumn{9}{l}{\textit{General Baselines}} \\
        MAGREF-480P~\cite{deng2025magref}   & 52.51 & 45.02 & 93.17 & \underline{21.81} & 30.83 & 70.47 & 43.04 & 66.90 \\
        VACE-P1.3B~\cite{jiang2025vace}     & 48.98 & \underline{47.34} & 96.80 & 12.03 & 16.59 & \textbf{71.38} & 40.19 & 64.31 \\
        VACE-1.3B~\cite{jiang2025vace}      & 49.89 & \textbf{48.24} & \underline{97.20} & 18.83 & 20.57 & \underline{71.26} & 37.91 & 65.46 \\
        VACE-14B~\cite{jiang2025vace}       & 57.55 & 47.21 & 94.97 & 15.02 & \textbf{55.09} & 67.27 & \underline{44.08} & 67.04 \\
        SkyReels-A2~\cite{fei2025skyreels}  & 52.25 & 39.41 & 87.93 & \textbf{25.60} & 45.95 & 64.54 & 43.75 & 60.32 \\
        Phantom-1.3B~\cite{phantom}         & 54.89 & 46.67 & 93.30 & 14.29 & 48.56 & 69.43 & 42.48 & 62.50 \\
        BindWeave~\cite{li2025bindweave}    & {57.61} & 45.55 & 95.90 & 13.91 & \underline{53.71} & 67.79 & \textbf{46.84} & 66.85 \\
        VINO~\cite{yuan2025opens2v}         & \underline{57.85} & 45.92 & 94.73 & 12.30 & 52.00 & 69.69 & 42.67 & \underline{71.99} \\
        SkyReels-v3~\cite{li2026skyreels}   & \textbf{59.26} & 46.92 & \textbf{99.78} & 15.17 & 45.90 & 67.98 & 39.70 & \textbf{84.12} \\
        \midrule

        \rowcolor{gray!10}\multicolumn{9}{l}{\textit{Results on Kaleido backbone}}\\
        Kaleido~\cite{zhang2025kaleido}     & 56.00 & \textbf{51.75} & \textbf{97.98} & 8.50  & 31.81 & \underline{70.37} & 38.62 & \underline{79.77} \\
        \quad + \method~($\gamma=0$)        & 56.32 & 50.40 & 97.07 & \textbf{10.05} & 31.79 & \textbf{70.54} & 41.22 & \textbf{79.86} \\
        \quad + \method~($\gamma=2$)        & 58.38 & 50.42 & \underline{97.09} & 9.84  & 43.80 & 70.23 & 42.18 & 77.78 \\
        \quad + \method~($\gamma=4$)        & {59.88} & \underline{50.45} & 97.02 & 9.94  & {52.37} & 69.92 & \underline{42.40} & {76.85} \\
        \quad + \method~($\gamma=8$)        & \underline{60.78} & 50.23 & 96.99 & \underline{9.96}  & \underline{57.46} & 69.65 & \underline{42.40} & 76.62 \\
        \quad + \method~($\gamma=16$)        & \textbf{61.14} & 50.11 & {96.70} & 9.71  & \textbf{60.64} & 69.50 & \textbf{43.02} & 75.23 \\
        \midrule

        \rowcolor{gray!10}\multicolumn{9}{l}{\textit{Results on Phantom-14B backbone}}\\
        Phantom-14B~\cite{phantom}          & 60.43 & \textbf{50.46} & 98.24 & \textbf{9.40} & 56.50 & \textbf{69.46} & 41.10 & \textbf{76.76} \\
        \quad + \method~($\gamma=0$)        & 60.41 & 50.29 & 98.39 & \underline{8.41} & 57.04 & \underline{69.14} & 42.76 & 75.14 \\
        \quad + \method~($\gamma=2$)        & 60.69 & 50.28 & 98.31 & 8.30 & 57.82 & \underline{69.14} & 42.81 & \underline{75.65} \\
        \quad + \method~($\gamma=4$)        & 60.62 & 50.29 & \underline{98.45} & 8.14 & 58.00 & 69.13 & 42.80 & 75.19 \\
        \quad + \method~($\gamma=8$)       & \underline{60.70} & \underline{50.32} & \textbf{98.47} & 8.33 & \underline{58.61} & 69.07 & \underline{42.83} & 75.00 \\
        \quad + \method~($\gamma=16$)        & \textbf{60.78} & 50.27 & 98.41 & 8.40 & \textbf{58.92} & 69.05 & \textbf{43.10} & 74.91 \\

        \bottomrule
    \end{tabular*}
\end{table}

\subsubsection{Quantitative Results}
\label{sec:quant_results}

As summarized in Table \ref{baselines_comparison}, \method achieves state-of-the-art performance on OpenS2V-Eval benchmark. A salient feature of our framework is its \textbf{controllable fidelity}, where the scale factor $\gamma$ enables monotonic improvement in identity preservation. Specifically, when integrated with the Kaleido backbone, increasing $\gamma$ from 2 to 8 yields a substantial rise in FaceSim from 43.80\% to 57.46\%, representing a 31.2\% relative gain without significant degradation in video naturalness. This scaling behavior underscores the efficacy of our Phase I step-wise guidance mechanism in anchoring subject-specific details within the denoising trajectory.
Crucially, the impact of our \textbf{Phase II RL anchoring} in mitigating \textbf{semantic drift} is empirically validated by the performance of \method at $\gamma=0$, where Phase I guidance is disabled. In this configuration, we observe a consistent improvement in \textbf{NexusScore}—a metric within the OpenS2V-Eval protocol designed to evaluate subject consistency and identity persistence. Compared to original backbones, \method ($\gamma=0$) improves NexusScore from 38.62\% to 41.22\% for Kaleido and from 41.10\% to 42.76\% for Phantom-14B. These gains directly link preference optimization to the reduction of semantic drift, confirming that Phase II effectively anchors the model's spatial attention to reference subjects during high-noise stages.

\subsubsection{Qualitative Results}
\label{sec:qual_results}
\begin{figure}[tb]
    \centering
    \includegraphics[width=1\linewidth]{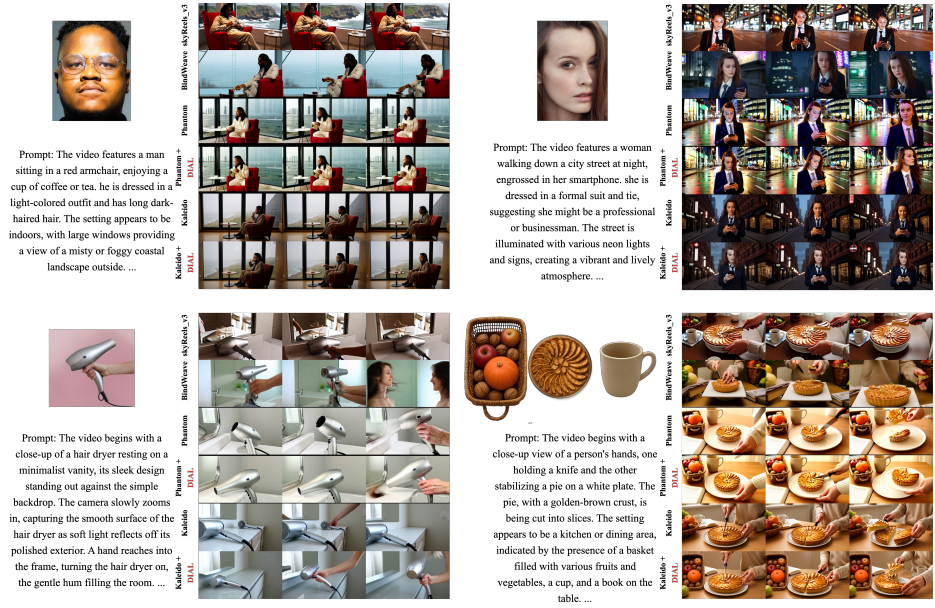}
    \caption{Qualitative comparisons with recent open-sourced models and our baselines.}
    \label{fig:visual_compare}
\end{figure}
As shown in \cref{fig:visual_compare}, \method exhibits stronger subject faithfulness and reduced semantic drift compared to SkyReels-v3 and BindWeave. It preserves fine-grained identity cues (e.g., hairstyle) and object attributes (e.g., hair dryer design) that baselines often corrupt. In multi-subject scenarios, \method prevents identity entanglement and geometric mismatches (e.g., fruit basket shape) seen in competitors. Compared to the backbone, \method enhances facial consistency and shape stability without sacrificing visual aesthetics or naturalness.

\subsection{Ablation Study}
\label{sec:ablation}

\subsubsection{Effect of \map in Fidelity Strength Control}
Fidelity strength control can be implemented without the use of \map by substituting the spatially-aware $\mathbf{S}_{i-1}$ in Eq. \ref{eq:mask} with a uniform all-ones matrix, a baseline variant we denote as Uni~\cite{tan2025ominicontrol}. As demonstrated in Table \ref{tab:ablation_enhance}, the Uni method suffers from a significant decline in NaturalScore even at a minimal enhancement scale of ($\gamma=2$). In contrast, \map ($\gamma=16$) achieves identity consistency (FaceSim) comparable to that of Uni ($\gamma=2$) while maintaining a much higher and more competitive NaturalScore of 75.23\%. These results indicate that applying a uniform strength enhancement across the entire latent space tends to degrade overall video realism, whereas \map provides localized, region-adaptive guidance that preserves visual quality. Additional qualitative comparisons supporting these findings are provided in Appendix \ref{app:stg1}.

\begin{table}[!t]
    \centering
    \footnotesize
    \caption{Ablation on Uni and \map in Fidelity Strength Control on the Kaleido backbone. \textbf{Bold} marks the best and \underline{underline} the second best value per column.}
    \label{tab:ablation_enhance}
    \begin{tabular*}{\linewidth}{@{\extracolsep{\fill}} l c c cc ccc c}
        \toprule
        \multirow{2}{*}{$\mathbf{S}_{i-1}$} & Total & Quality & \multicolumn{2}{c}{Dynamics} & \multicolumn{3}{c}{Consistency} & Realism \\
        \cmidrule{2-2} \cmidrule{3-3} \cmidrule{4-5} \cmidrule{6-8} \cmidrule{9-9}
        & Score$\uparrow$ & Aes.$\uparrow$ & Smth.$\uparrow$ & Amp.$\uparrow$ & Face$\uparrow$ & Gme.$\uparrow$ & Nex.$\uparrow$ & Nat.$\uparrow$ \\
        \midrule
        $\gamma=0$               & 56.32 & 50.40 & 97.07 & \textbf{10.05} & 31.79 & \textbf{70.54} & 41.22 & \textbf{79.86} \\
        \map~($\gamma=2$)        & 58.38 & \underline{50.42} & \underline{97.09} & 9.84  & 43.80 & \underline{70.23} & 42.18 & \underline{77.78} \\
        \map~($\gamma=4$)        & 59.88 & \textbf{50.45} & 97.02 & 9.94  & 52.37 & 69.92 & 42.40 & 76.85 \\
        \map~($\gamma=8$)        & \underline{60.78} & 50.23 & 96.99 & \underline{9.96}  & 57.46 & 69.65 & 42.40 & 76.62 \\
        \map~($\gamma=16$)       & \textbf{61.14} & 50.11 & 96.70 & 9.71  & \textbf{60.64} & 69.50 & \underline{43.02} & 75.23 \\
        Uni~($\gamma=2$)         & 59.63 & 50.23 & \textbf{97.39} & 9.75 & \underline{59.21} & 69.47 & \textbf{43.26} & 69.03 \\
        \bottomrule
    \end{tabular*}
\end{table}

\subsubsection{Effect of Multi-step DPO Iterations}
\begin{wrapfigure}[15]{r}{0.44\linewidth}
  \centering
  \includegraphics[width=\linewidth]{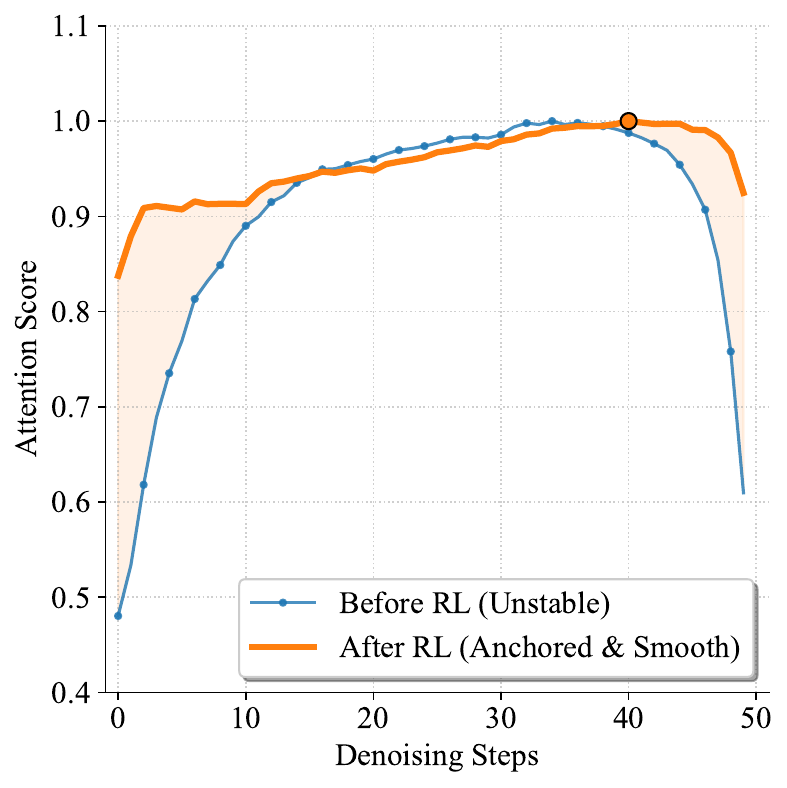}
  \caption{Evolution of attention scores (with normalization) across denoising steps.}
  \label{fig:attn_comparison}
\end{wrapfigure}

\begin{table}[!t]
    \centering
    \footnotesize
    \caption{Ablation on the number of DPO iterations ($T_{\mathrm{dpo}}$) on the Kaleido backbone with $\gamma{=}0$. $T_{\mathrm{dpo}}{=}0$ corresponds to the baseline. \textbf{Bold} marks the best value per column.}
    \label{tab:ablation_dpo}
    \begin{tabular*}{\linewidth}{@{\extracolsep{\fill}} l c c cc ccc c}
        \toprule
        \multirow{2}{*}{$T_{\mathrm{dpo}}$} & Total & Quality & \multicolumn{2}{c}{Dynamics} & \multicolumn{3}{c}{Consistency} & Realism \\
        \cmidrule{2-2} \cmidrule{3-3} \cmidrule{4-5} \cmidrule{6-8} \cmidrule{9-9}
        & Score$\uparrow$ & Aes.$\uparrow$ & Smth.$\uparrow$ & Amp.$\uparrow$ & Face$\uparrow$ & Gme.$\uparrow$ & Nex.$\uparrow$ & Nat.$\uparrow$ \\
        \midrule
        0 & 56.00 & \textbf{51.75} & 97.98 & 8.50  & \textbf{31.81} & 70.37 & 38.62 & 79.77 \\
        1 & 56.00 & 51.11 & \textbf{98.09} & 9.80 & 31.74 & 70.19 & 39.20 & 79.60 \\
        2 & \textbf{56.32} & 50.40 & 97.07 & 10.05 & 31.79 & \textbf{70.54} & \textbf{41.22} & \textbf{79.86} \\
        3 & 55.64 & 50.65 & 98.07 & \textbf{13.61} & 31.68 & 70.40 & 38.71 & 78.55 \\
        \bottomrule
    \end{tabular*}
\end{table}
We first study how the multi-step preference horizon, denoted by $T_{\mathrm{dpo}}$, affects Phase~II alignment. 
Table~\ref{tab:ablation_dpo} shows that $T_{\mathrm{dpo}}{=}2$ yields the best overall trade-off, improving identity-related metrics while maintaining stable realism.
With $T_{\mathrm{dpo}}{=}1$, the preference signal is computed from a single step and becomes less reliable, leading to weaker identity preservation.
Increasing to $T_{\mathrm{dpo}}{=}3$ brings marginal gains on some consistency metrics but slightly degrades overall realism, suggesting over-emphasis on early-step preferences.
We therefore use $T_{\mathrm{dpo}}{=}2$ as the default setting in all experiments.

\subsubsection{Attention Dynamics Enhancement in Phase II}
We examine how Phase~II changes reference grounding over denoising timesteps. Fig.~\ref{fig:attn_comparison} reports the average grounding score across timesteps. Before preference optimization, the score is low in the initial high-noise steps, matching our observation that early attention is diffuse. After Phase~II, the early-stage score increases markedly, and the curve becomes smoother, indicating that anchoring encourages correct reference attention earlier and reduces error propagation to later refinement.
We include more in-depth analysis in Appendix~\ref{app:stg2}.

\section{Conclusion}
We introduced \method, a unified dual-phase framework for controllable subject-to-video (S2V) generation. Central to our work is the identification of \mapfull (\map) within specific DiT blocks, which provide reliable and intrinsic subject localization signals. \method leverages these maps through a two-phase strategy: \textbf{Phase I} implements training-free, inference-time guidance for fine-grained and independent fidelity control; and \textbf{Phase II} utilizes a zero-cost preference optimization to anchor subject semantics during high-noise stages. Extensive experiments on the OpenS2V-Eval benchmark demonstrate that \method achieves state-of-the-art performance, significantly improving identity consistency and mitigating semantic drift without compromising visual quality.



%
%
\bibliographystyle{splncs04}
\bibliography{main}

\clearpage
\setcounter{section}{0}
\renewcommand\thesection{\Alph{section}}

\section{Additional Qualitative Results for Low-noise Inference Guidance}
\label{app:stg1}

Phase~I performs training-free attention guidance in the low-noise denoising regime by injecting an ISGM-conditioned bias into attention logits (cf.\ Eq.~\ref{eq:mask}--\ref{eq:phase1_attn} in the main paper).
Here we provide additional qualitative results to support three claims: (i) ISGM is crucial for \emph{localized} fidelity enhancement, (ii) the ISGM signal becomes reliable only after sufficient denoising, and (iii) the fidelity scale $\gamma$ offers smooth and robust controllability.

\begin{figure}[tb]
    \centering
    \includegraphics[width=1\linewidth]{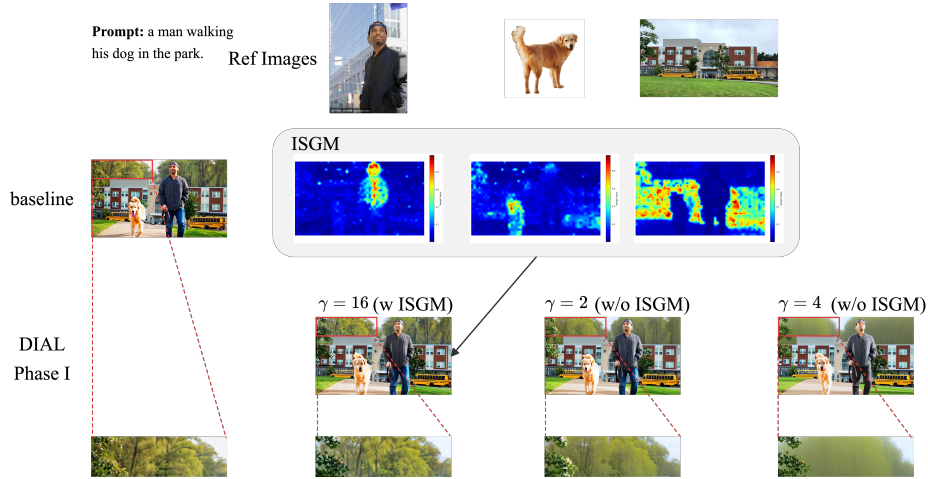}
    \caption{\textbf{Localized (ISGM-guided) vs.\ global guidance in Phase~I.}
    We compare our ISGM-based spatially localized guidance with a global baseline that applies a uniform guidance signal to all spatial locations.
    Without ISGM, increasing $\gamma$ noticeably degrades overall video quality; the magnified insets show severe background blurring and texture loss already at $\gamma=2$ and $\gamma=4$.
    In contrast, \textbf{ISGM-guided} attention bias allows strong fidelity enhancement (e.g., $\gamma=16$) while preserving sharp backgrounds, indicating that our guidance is confined to subject-specific regions.}
    \label{fig:stg1_demo_baseline}
\end{figure}

\paragraph{ISGM enables localized enhancement.}
We first compare ISGM-guided attention bias with a global baseline that applies uniform guidance to all spatial locations.
As shown in Fig.~\ref{fig:stg1_demo_baseline}, uniform guidance tends to introduce global artifacts, whereas ISGM guidance preserves the background while strengthening subject fidelity.

\begin{figure}[tb]
    \centering
    \includegraphics[width=1\linewidth]{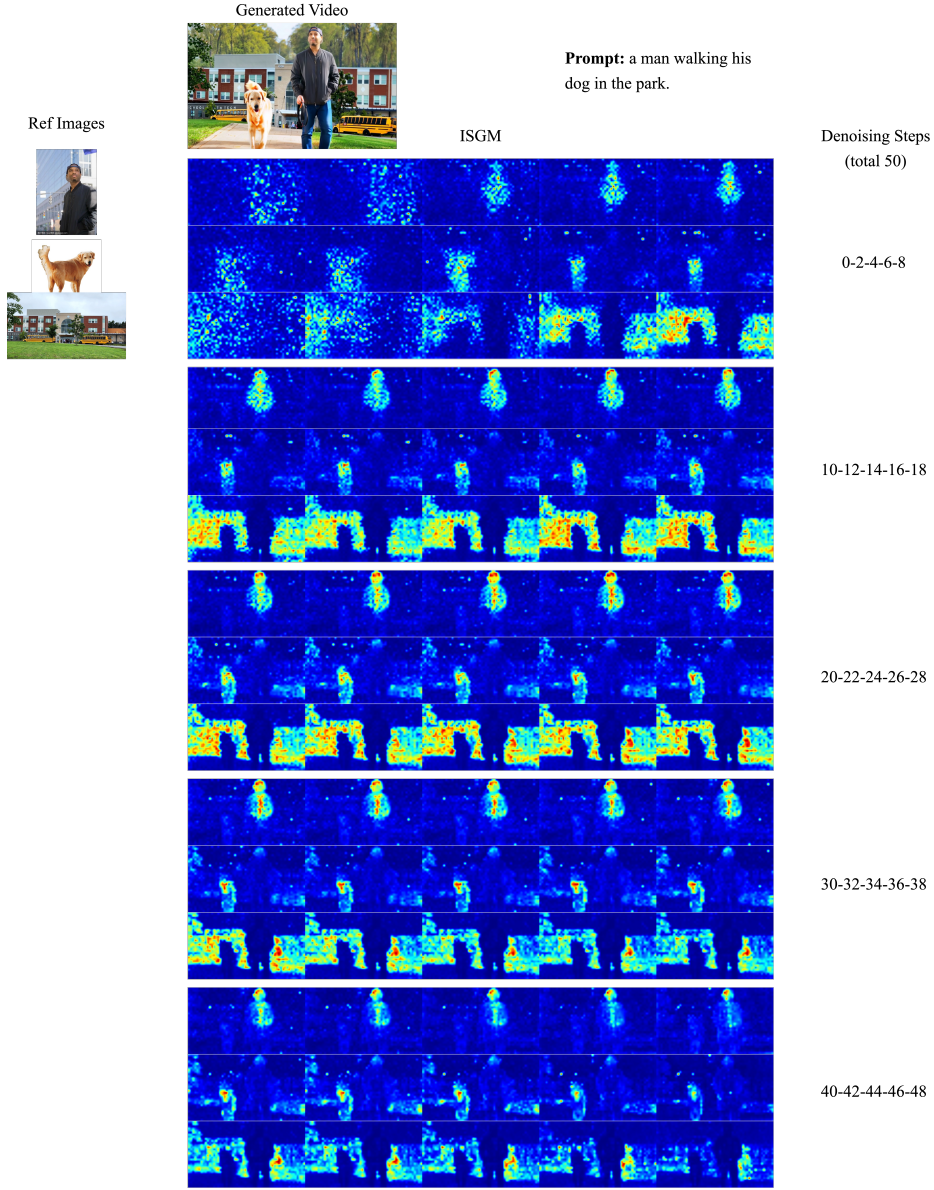}
    \caption{\textbf{Evolution of ISGM across denoising steps.}
    In the early high-noise regime (roughly the first 20\% steps; e.g., steps 0--8), ISGM is diffuse and provides unreliable spatial localization; thus we disable Phase~I guidance in this regime.
    As denoising proceeds, ISGM becomes progressively sharper and expands to cover the full subject extent.
    Near the final low-noise steps (e.g., $>40$), the activation often concentrates on discriminative regions (e.g., faces), which are critical for identity preservation.}
    \label{fig:isgm_over_steps}
\end{figure}

\paragraph{ISGM is reliable mainly in the low-noise regime.}
Fig.~\ref{fig:isgm_over_steps} visualizes how ISGM evolves across denoising steps.
The early high-noise stage yields scattered maps, motivating our choice to disable Phase~I guidance there; as denoising proceeds, the maps become sharply localized, enabling effective spatially conditioned control.

\begin{figure}[tb]
    \centering
    \includegraphics[width=1\linewidth]{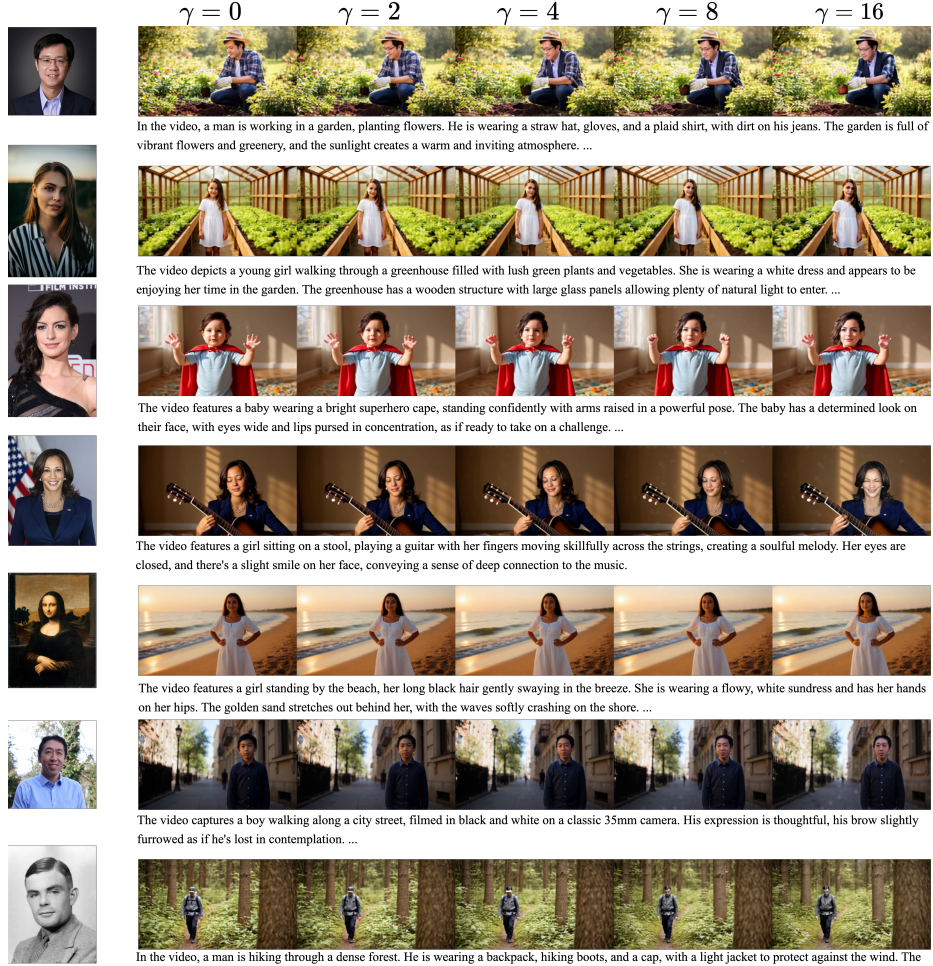}
    \caption{\textbf{Effect of fidelity scale $\gamma$ on identity preservation.}
    Increasing $\gamma$ from 0 to 16 yields a monotonic improvement in identity similarity.
    This controllability remains effective under challenging conditions, including \textbf{prompt--reference ambiguity} (Row~3: prompt ``a baby'' with an adult reference) and \textbf{cross-domain generation} (Row~5: reference image in an oil-painting style).
    The backbone is Kaleido~\cite{zhang2025kaleido}.}
    \label{fig:demo_for_strength}
\end{figure}

\paragraph{Continuous controllability via $\gamma$.}
Fig.~\ref{fig:demo_for_strength} illustrates that increasing $\gamma$ yields consistent improvements in identity similarity on Kaleido~\cite{zhang2025kaleido}, and remains effective under prompt--reference ambiguity and cross-domain references.

\section{Analysis of Semantic Drift Mitigation}
\label{app:stg2}

Phase~II targets the high-noise regime where deterministic ISGM-based bias injection is less reliable.
We therefore anchor reference grounding via preference optimization over denoising trajectories (cf.\ Algorithm~\ref{alg:loss} in the main paper).

\paragraph{Quantifying subject disappearance.}
We quantify subject presence using an image-prompt open-vocabulary detector~\cite{cheng2024yolo}, with each reference image as the query.
We use the detector implementation released with OpenS2V-Eval \footnote{YOLO-World checkpoint: \nolinkurl{yolo_world_v2_l_image_prompt_adapter-719a7afb.pth}.}.
As shown in Table~\ref{tab:det_improve_for_RL}, Phase~II increases the match ratio, indicating that fewer reference subjects are completely missing from the generated videos (with Phase~I disabled, $\gamma=0$).

\begin{table}[tb]
    \centering
    \footnotesize
    \caption{\textbf{Phase~II improves subject presence (video-level detection).}
    We evaluate subject presence using an image-prompt open-vocabulary detector~\cite{cheng2024yolo}, where each reference image is used as the query.
    The benchmark contains 180 cases with varying numbers of reference subjects, resulting in 320 \emph{(case, subject)} instances in total.
    An instance is counted as a \textbf{Match} if the queried subject is detected in \emph{at least one frame} of the generated video; otherwise it is counted as a miss.
    ``Total'' is the number of evaluated (case, subject) instances, and ``Ratio'' is computed as Matches/Total.
    All results are reported with Phase~I guidance disabled ($\gamma=0$) to isolate the effect of Phase~II.}
    \label{tab:det_improve_for_RL}
    \begin{tabular*}{\linewidth}{@{\extracolsep{\fill}} l c c c}
        \toprule
        Method & Matches$\uparrow$ & Total & Ratio$\uparrow$ \\
        \midrule
        Kaleido~\cite{zhang2025kaleido} & 155 & 320 & 0.48 \\
        \quad + \method~($\gamma=0$) & 163 & 320 & 0.51 \\
        \bottomrule
    \end{tabular*}
\end{table}

\begin{figure}[tb]
    \centering
    \includegraphics[width=1\linewidth]{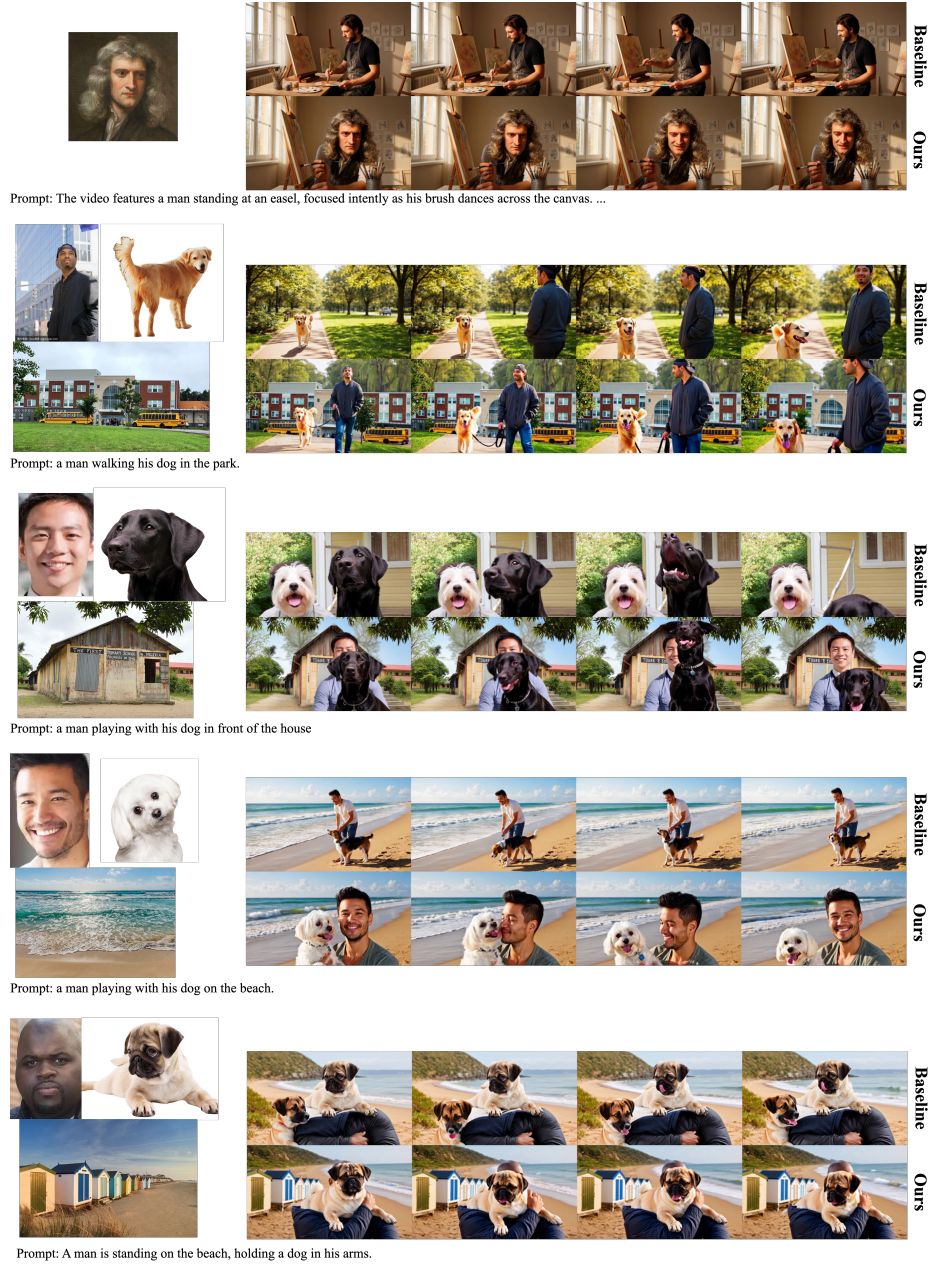}
    \caption{\textbf{Effect of Phase~II high-noise anchoring via preference optimization.}
    \textbf{Baseline:} Kaleido backbone~\cite{zhang2025kaleido}.
    \textbf{Ours:} Phase~II model with Phase~I guidance disabled ($\gamma=0$).}
    \label{fig:demo_for_rl}
\end{figure}

\paragraph{Qualitative evidence.}
Fig.~\ref{fig:demo_for_rl} shows representative examples where Phase~II better preserves the identities and key attributes of the reference subjects than the Kaleido backbone, consistent with the reduced subject disappearance reported in Table~\ref{tab:det_improve_for_RL}.

\section{Additional Implementation Details}
\label{app:exp}

\subsection{Identification of Intrinsic Spatial Grounding Maps (ISGM)}
\paragraph{Attention score.}
In Fig.~\ref{fig:isgm_peak}, we quantify the grounding strength of an attention block by the ratio between attention mass inside the predicted subject region and that outside.
For each reference $k\in\{1,\dots,K\}$, we compute
\begin{equation}
r_k \;=\;
\frac{
\frac{1}{|\mathcal{I}^{k}_{+}|}\sum_{q\in\mathcal{I}^{k}_{+}} s_k(q)
}{
\frac{1}{|\mathcal{I}^{k}_{-}|}\sum_{q\in\mathcal{I}^{k}_{-}} s_k(q)
},
\label{eq:rk_ratio}
\end{equation}
where $q$ indexes video queries (patch tokens), $s_k(q)$ is the per-reference attention score at query $q$ (cf.\ Eq.~(5) in the main paper), and $\mathcal{I}^{k}_{+}$ / $\mathcal{I}^{k}_{-}$ denote the foreground/background query sets, respectively.
We then aggregate across references as
\begin{equation}
R \;=\; \frac{1}{K}\sum_{k=1}^{K} r_k .
\label{eq:R_avg}
\end{equation}

\paragraph{Foreground/background region construction.}
We use the image-prompt detection model provided by OpenS2V-Eval~\cite{yuan2025opens2v} to obtain a bounding box for each reference subject in each generated frame.
We map the predicted box to the latent token grid to form $\mathcal{I}^{k}_{+}$, and define $\mathcal{I}^{k}_{-}$ as its complement over video query tokens.
The score reported in Fig.~\ref{fig:isgm_peak} is the average of $R$ over all 180 prompts in OpenS2V-Eval.

\subsection{\method Phase~I: Low-noise Inference Guidance}
\paragraph{ISGM computation.}
In practice, ISGM is extracted from a pre-identified privileged attention block (see Fig.~\ref{fig:isgm_peak}).
We compute ISGM from this block independently at each denoising step.
Since only a single block is used, the additional overhead is minimal.

\paragraph{Inference settings.}
Some baselines adopt prompt rephrasing during OpenS2V evaluation.
For Kaleido~\cite{zhang2025kaleido}, we use the original prompts; for Phantom~\cite{phantom}, we follow the official recommendation to rephrase prompts before evaluation.
Unless specified otherwise, all evaluations use $T=50$ denoising steps and the default resolution of each backbone.

\subsection{\method Phase~II: High-noise Anchoring via Preference Optimization}
\paragraph{Training set.}
We sample 14K multi-reference training examples from Phantom-Data~\cite{chen2025phantom}.
We use SAM3~\cite{carion2025sam} to obtain pseudo segmentation masks $\{\mathbf{M}_k\}_{k=1}^{K}$ for each subject as supervision for preference construction.
Although Phantom-Data has limited background-style references, we observe that Phase~II training still improves consistency for background references (see Fig.~\ref{fig:demo_for_rl}), suggesting that our anchoring objective improves reference utilization and generalization.

\paragraph{Training hyperparameters.}
We train with 48 NVIDIA H100 GPUs using LoRA with rank 16.
We use a learning rate of $1\times10^{-4}$ and train for 800 steps (approximately two epochs).

\end{document}